# Improving Accuracy and Scalability of the PC Algorithm by Maximizing P-value[1]

Joseph Ramsey
Technical Report
Center for Causal Discovery
October 2, 2016

**Abstract**

**A number of attempts have been made to improve accuracy and/or scalability of the PC (Peter and Clark) algorithm, some well known (Buhlmann, et al., 2010; Kalisch and Buhlmann, 2007; 2008; Zhang, 2012, to give some examples). We add here one more tool to the toolbox: the simple observation that if one is forced to choose between a variety of possible separating sets for a pair of variables, one should choose the one with the highest p-value. One can use the CPC (Conservative PC, Ramsey et al., 2012) algorithm as a guide to possible separating sets for a pair of variables. However, whereas CPC uses a voting rule to classify colliders versus noncolliders, our proposed algorithm, PC-Max, picks the conditioning set with the highest p-value, so that there are no ambiguities. We combine this with two other optimizations: (a) avoiding bidirected edges in the orientation of colliders, and (b) parallelization. For (b) we borrow ideas from the PC-Stable algorithm (Colombo and Maathuis, 2014). The result is an algorithm that scales quite well both in terms of accuracy and time, with no risk of bidirected edges.**

**1. Maximum P-value.**

Consider the following problem. You know that X and Y are independent conditional on some set of variables S, but you don't know which. You have a candidate list; it could be any one of $S_1, S_2,..., S_m$. You do an independence test for X and Y conditional on each of these subsets and get p-values $p_1, p_2, ..., p_m$. Based on this information alone, which set S would you pick, if you wanted to make sure that X and Y were most likely not to be dependent conditional on S?

The answer that suggests itself is that you should pick the one with the *highest p-value*; this is the one that is least likely to be dependent, in an appropriate sense. A simple argument can be made.

First of all, under the null hypothesis $H_0$ of independence, after transforming partial correlations to Z values using the Fisher Z transform, which we can always do for linear, Gaussian models, $Z \sim N(0, 1)$. This distribution depicts how values of Z distribute under $H_0$. All hypotheses that it depicts are independencies. So you could ask for this distribution, how can one maximize or minimize the number of independencies found, in probability? The answer is, by making alpha larger or smaller, since P(dependence) = P(|z| > alpha | $H_0$), where z is the Fisher Z value for a particular test.

That's all well and good, but it's not the question just asked. What we want to know is how we can make the probability of *dependence* larger or smaller. For that, we need to consider the alternative hypothesis, $H_1$, of dependence. This is a difficult distribution to nail down. In fact, to get it, one may need to merge together distributions from different specific alternative hypotheses, which sets a high bar for success. Nevertheless in any particular case, there is such a distribution for Z under $H_1$. Then on the usual way of judging dependence, if you pick an alpha, P(dependence) = P(|z| < alpha | $H_1$). To make this probability larger, raise alpha; to make it smaller, lower alpha. In this way, you can directly control the probability of a positive dependence judgment, and hence in probability at least, number of positive dependence judgments.

[1] Research supported by the National Institutes of Health under Award Number U54HG008540. The content is solely the responsibility of the author and does not necessarily represent the official views of the National Institutes of Health. Thanks to Clark Glymour for helpful comments on previous drafts.

To illustrate, Hung et al. (1997) make a similar point in a less general context by giving the distribution of p-values for Z (calculated under H0) under $H_1$, where under H0 (Z = 0) Z is distributed as N(0, 1), and under $H_1$ Z (under their assumption Z > 0) is distributed as

$$g_\delta(p) = \phi(Z_p - \sqrt{n}\delta)/\phi(Z_p), \qquad 0 < p < 1,$$

where $\delta = \mu / \sigma$, where the alternative hypothesis is assumed to be distributed as N($\mu$, $\sigma^2$), and $\varphi$ is the Normal cumulative distribution function. Plotted, one gets curves as in Figure 1.

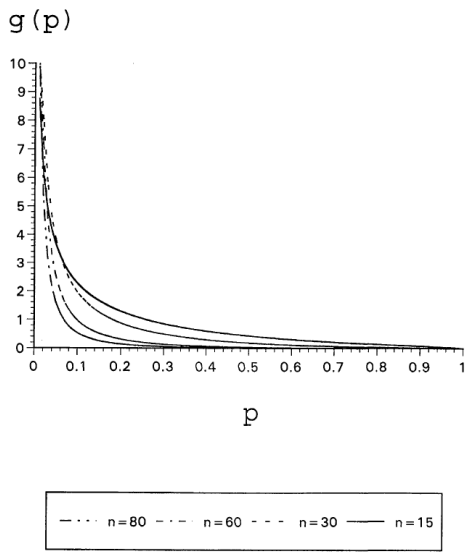

**Figure 1. Density of p-value for various n at delta = 1/3. Figure borrowed from Hung et al. (1997).**

What's clear from these plots, as well as from g, is that, at least for fixed $\mu$ and $\sigma$, the distribution of p-values is strongly weighted to values near 0 and declines monotonically from there up to 1. Thus, if you have a list of p-values for tests and want to pick the one with the lowest likelihood of dependence, as judged by the value of g, you should pick the one furthest to the right—that is, the maximum p-value.

That's the argument; no more is needed. Notice that we don't even need to raise the question of whether a judgment of dependence is *false*. We simply find the Z value for which the region judged dependent is the smallest, if we were to use that Z value as our alpha. To illustrate, if we choose an alpha for a two-tailed test, say, then probability P(dependence) = P(|Z| < alpha | $H_1$)--that is, the probability that a judgment with |Z| < alpha is for dependence. In that case, P(dependence) = P(|Z| < z | $H_1$), where z is the Z value for a particular test is the minimum of P(|Z| < alpha | $H_1$) for alpha > z. Then z is the minimum alpha you could use while still judging z to be a Z value for a test of dependence

Happily, this sort of reasoning about $H_1$ does not depend on linearity or Gaussianity at all; it works for any test, so long as we distribute the same statistic under both $H_0$ and $H_1$. We don't care what the distribution of our statistic is under $H_1$, so long as there is one. That is, we should pick the test statistic from those calculated by whatever test we use that *minimizes* the relevant dependence region under the distribution of $H_1$. But that is to say, we should *maximize* the p-value (area in the tails) that we calculate under $H_0$.

Similar reasoning can be carried out for scores. Consider, for instance, the linear, Gaussian BIC score. Chickering (2002) points out that one may estimate conditional independence in the linear, Gaussian case as a difference of two BIC scores. We may use the formula for BIC (different formulas are given, up to scaling), assuming the data are i.i.d., as follows:

BIC = 2 L – k ln N

where L is the likelihood, k is the number of degrees of freedom, and N is the sample size. To estimate conditional independence of X and Y conditional on $Z_1,..,Z_n$, form this model:

X->Y<-{$Z_1,...,Z_n$}

and estimate its BIC score $B_1$, then form this model:

Y<-{$Z_1,...,Z_n$}

and estimate its BIC score $B_2$, then simply judge independence if $B_2 – B_1 < 0$. To put it another way, we judge independence if adding the variable X to the model makes no difference to the likelihood; the difference may be negative because $B_1$ has one more degree of freedom than $B_2$. One calculates the likelihood for one of these models by regressing Y onto its parents and calculating the residual $e_Y$, which is Normally distributed. One calculates the likelihood, therefore, by summing the log probabilities of the Normal density function for each data point. Usually for this likelihood we substitute the sample variance for the population variance and get

L = -n / 2 ln $s_1^2$

But if we don't do this substitution, we get

B1 = -n / 2 ln $\sigma_1^2$ – n / 2 ($s_1^2 / \sigma_1^2$) – (k + 1) ln N < 0

We may do the same for B2:

B2 = -n / 2 ln $\sigma_2^2$ – n / 2 ($s_2^2 / \sigma_2^2$) – k ln N < 0

But this is true just in case for some C (calculated easily from the constants in the above equations),

($s_1^2 / \sigma_1^2$) / ($s_2^2 / \sigma_2^2$) < C

This statistic is distributed as F. So if one is faced with the problem of choosing a set S such that X _||_ Y | S, and one has several such sets to choose from, the likelihood of each one being less than C will depend on the area to the left of this statistic, so if one wants to choose the score with the least likelihood of rendering X and Y dependent, one should choose the one such that B2 – B1 is minimal, that is, the one with the highest p-value under the F with the appropriate degrees of freedom. Similar transformations can be made for other types of scores. The choice of testing versus scoring is heuristic; they simply rely on different distributions. In general, scoring is faster, because computationally the cumulative distribution function for the Normal distribution doesn't have to be called as often, though usually one can finesse this difference with testing by precalculating Z cutoffs.

How, then, can we use this maximum p-value idea in an algorithm? We will put together several ideas to produce our final working algorithm, though the centerpiece will be this idea of maximizing p-value. We will use ideas from the PC algorithm (Spirtes et al., 1993), the Conservative PC algorithm (Ramsey et al., 2012) and the PC-Stable algorithm (Colombo and Maathuis, 2014), together with the maximum p-value notion, and we will parallelize the algorithm where possible to speed it up. That will be the final proposed algorithm, which we will compare to some of the existing algorithms in simulation. We will show that the algorithm scales in time easily, up to a 20,000 variables for sparse graphs on a laptop,[2] and quite accurately as well, far more accurately than any of the compared constraint-based competitors. But to get there we have to build up the argument piece by piece. We start with the well-known algorithm PC (Peter and Clark, Spirtes et al., 2000) and show how the maximum p-value idea can be incorporated. We will also suggest two other modifications that will help. Specifically, we will modify CPC (Conservative PC, Ramsey et al., 2012), which has already the apparatus to make the maximum p-value method work, with a simple modification. We begin by reviewing the PC and CPC algorithms.

**2. PC.**

Pseudocode for the PC algorithm is given in Figure 2. $I(X, Y | Z_1,...,Z_n)$ returns true if $X \perp\!\!\!\perp Y | Z_1,...,Z_n$ is judged true, otherwise false. V is the set of variables being searched over. It is assumed that all common causes of variables in V are also in V (causal sufficiency) and that the graph for the true causal model is directed and acyclic (a DAG).

**Figure 2. Pseudocode for the PC algorithm. See text.**

Procedure PC(I, V)
1. G <- complete graph over V
2. m <- 0
3. S:edges->sets of variables.
4. Repeat until no more changes can be made
    a. Remove from G all edges X—Y that are independent conditional on m variables from adj(X) or adj(Y)
    b. Map these removed edges to the sets used to remove them, in S.
    c. m <- m + 1
5. For each unshielded triple X—Y—Z in G, orient this as X->Y<-Z if S(X—Z) does not contain Y.
6. Use the rules in Meek (1995) to orient any further edges in G that can be oriented without create new unshielded colliders
7. Return G

The algorithm consists of an adjacency phase followed by an orientation phase. In the adjacency phase, a complete graph G over a set of variables V is constructed, and then edges X—Z are removed from G if a conditioning set S is found such that $X \perp\!\!\!\perp Z | S$. The algorithm looks for these conditioning sets in a certain order, as indicated in the pseudocode, for reasons of efficiency. Then, in the orientation phase, unshielded triples are located in G, that is graph snippets of the form X—Y—Z, with X and Z not adjacent (unshielded triples), and these are oriented as *colliders*, X->Y<-Z, just in case the conditioning set that was used to remove the X—Z edge in the adjacency phase does *not* contain Y. Once all of the unshielded colliders have been oriented in this fashion, as many further orientations are made as possible without creating new unshielded colliders. The rules that are used to accomplish this are given in (Meek, 1995).

**3. Conservative PC.**

One problem with the PC algorithm as written is that if judgments of conditional independence are not implied by judgments of conditional d-separation in the true model, or vice-versa, it's possible for collider judgments to be mistaken, and it's also possible for the final orientation rules to be applied illegitimately. So for example the rule R1 can orient long chains of directed edges, where these chains should have been stopped early on because some particular unshielded triple was not in fact a noncollider. The Conservative PC (CPC) algorithm tries to address this, using a *unanimous voting rule*, as follows. We know (Spirtes et al., 1993) that if X and Z are conditionally independent, then there is some set S either in adj(X) or adj(Z) such that X _||_ Z | S. So we can simply test all of these different sets for conditional independence. For orienting an unshielded triple X—Y—Z as a collider, we simply want to test whether S contains Y; if it doesn't, we orient the triple as X->Y<-Z. Notice that the PC algorithm will pick one particular such set in the course of its operation, but it's not necessary to look at just one set; we could look at all of the subsets of the adjacent variables of X or of Z. In fact, we could make a list of all of the subsets S of adj(X) or adj(Z) such that X _||_ Z | S. Now if conditional independence facts corresponded to conditional d-separation facts, if would be the case either that all of these sets contained Y or that all of them did not contain Y; there would be no ambiguity. And yet there can be ambiguity for small samples. The CPC algorithm (Ramsey et al., 2016) marks such ambiguities in the output graph with underlines; see the pseudocode in Figure 3. The way the CPC algorithm modifies the PC algorithm is just as described. The adjacency phase is unchanged, except that conditioning sets are not stored. Rather, for orienting colliders, conditioning sets yielding independence ("sepsets") for X and Z are sought from among adj(X) or adj(Z). These are listed, and it is determined whether a unanimous judgment can be made as to whether for a unshielded triple X—Y—Z Y is in all of these sepsets or in none of them. If in none, the unshielded triple is oriented as a collider; if in all, it is left a noncollider; if in some but not all, it is judged ambiguous and underlined.

---

[2] It is clear from Tables 1 and 2 that PC-Max can scale accurately and quickly to much larger problems; we will try larger problems on a supercomputer as time permits. All of the runs reported in these tables were done on a laptop with four processors, as described.

Pseudocode for the CPC algorithm is given in Figure 3.

**Figure 3. Pseudocode for the CPC algorithm.**

Procedure CPC(I, V)
1. G <- complete graph over V
2. m <- 0
3. Repeat until no more changes can be made
    a. Remove from G all edge X—Y that are independent conditional on m variables from adj(X) or adj(Y)
    b. m <- m + 1
4. For each unshielded triple X—Y—Z in G, determine for which subsets X of adj(X) or adj(Z) it is the case that I(X, Z | S); list these subsets as L.
5. If Y is not in any subset in L, orient X->Y<-Z in G; if Y is in all subsets of L, leave the triple unoriented in G (this will be a noncollider); otherwise, underline X—Y—Z in G to mark it as ambiguous.
6. Use the rules in Meek (1995) to orient any further edges that can be oriented in G without create new unshielded colliders, respecting the underlinings in the graph.
7. Return G

The upshot is that CPC makes fewer orientations than PC, but the orientations it makes are more secure. It can still orient bidirected edges, but in general this is good news. The downside is that for real data it has been reported to make very few orientations for biological data (Maathuis, personal communication), so the conservativeness may be unhelpful for making positive orientations.

**4. Replacing Conservative Orientation with Max P-value Orientation**

Note that the conservative orientation of colliders can now be replaced quite directly by the maximum p-value method. That is, when orienting an unshielded triple X—Y—Z, one can list the various subsets $S_1,...,S_n$ of adj(X) or adj(Z) and calculate the Fisher Z transform of the partial correlation for each independence judgment, yielding p-values $p_1,...,p_n$. The goal is to choose such a subset that yields independence, and as in Section 1, we should choose the one with the highest p-value. We make this modification.

A premise of this method is that conditioning on one of $S_1,...,S_n$ yields an independence. This is justified since the adjacency search has taken out the edge X—Z. So X and Z must be independent conditional on some set, and a theorem in Spirtes et al. (2000) states that if they are independent conditional on some set, then they must be independent conditional on some subset of adj(X) or adj(Y). This may be false in small samples, but it is correct in the large sample limit under the assumptions of the PC algorithm—namely, that there are no unrecorded common causes, that the true causal model is a directed acyclic graph (DAG), and that the conditional independence test used is correct in the large sample limit. This finesses in a sense the problem of *orientation unfaithfulness* addressed by CPC (Ramsey et al., 2012); it sets that problem aside and provides an answer in the orientation phase to every conditional independence question posed. The reason this works is that when faced with a choice of different conditioning sets on which X and Z might be independent, it simply chooses the one that is least likely to be an actual judgment of dependence, in the sense explained in Section 1. This is all that's necessary for the collider orientation step of PC; we simply need to find one true positive sepset for each unshielded triple. We don't care if there is more than one; we just care that there is one from this set of alternatives.

As noted above, if we make this modification, there are no longer any ambiguities in our output; all unshielded triples are definitively classified as either colliders or noncolliders. We are in essence choosing a *best* sepset in each case, instead of using a voting rule. The adjacency phase tells us, when X is not adjacent to Z, that X and Z are independent conditional on some set of variables. We simply use this rule to determine the best candidate for such a set. In the large sample limit it's correct, since in the large sample limit the adjacency phase will not make a mistake; there will be a sepset from among adj(X) or adj(Z) which in the large sample limit will be judged independent. We could pick any such subset yielding independence; we simply prefer the one with the highest p-value, since that's the one that's least likely to create a dependency.

That's the proposal. How well does it work? Before addressing that, we first address another thorny issue, bidirected edges.

**5. Avoiding Bidirected Edges.**

One problem with the PC algorithm, as well as the CPC algorithm modified as in Section 5, to an extent, is that it's possible to orient bidirected edges, which can't exist in the correct output of the algorithm (a pattern or CP-DAG). The mechanism for this is completely obvious and unavoidable, unless some correction is made. Say you have two unshielded triples connected together, X—Y—Z—W. If you happen to orient both X—Y—Z and Y—Z—W as colliders, because of small sample errors, or false positive adjacencies in the graph, or because latent common causes exist in your dataset, then you will end up with X->Y<->Z<-W, and you will have a bidirected edge Y<->Z. You can only avoid this if you avoid orienting both colliders at the same time. But which one should you orient? We propose, by the same max p-value reasoning proposed in section 1, to orient the collider whose conditioning set yields the higher p-value. In fact, for systematic reasons, we propose to precalculate the p-values for all colliders that could be added to the graph, then to sort these high to low, and then to orient their corresponding colliders one by one in the graph, refusing to orient any collider in the process that could create a bidirected edge. This is correct in the large sample limit, if the assumptions of the PC algorithm hold, since the calculated colliders cannot conflict.

The obvious problem with this method of avoiding bidirected edges is that one may make mistakes, and if one does, there will be edges oriented in directions they shouldn't be. So will end up with a pattern but a pattern in which some edges are oriented in the wrong direction. But this is not an issue of correctness, since in the large sample limit this can't happen.

## 6. The PC-Max Algorithm.

Pseudocode for the PC-Max algorithm is given in Figure 4. S(X, Y | Z) is a function that returns the p-value of the independence test for X _||_ Y | Z. For the scoring interpretation, this will be maximized when the score is minimized.

### Figure 4. Pseudocode for the CPC algorithms.

Procedure PC-Max(I, V, S)
1. G <- complete graph over V
2. m <- 0
3. Repeat until no more changes can be made
    a. Start a list of edges to be removed, E.
    b. For each edge X—Y in G
        i. If I(X, Y | S) for some set S of size m, where S subset of adj(X) or S subset of adj(Y)
            1. Remove X—Y from G.
    c. m <- m + 1
4. Start a new list C of collider triples, mapped to the p values.
5. For each unshielded triple X—Y—Z in G
    a. Estimate independence each subsets X of adj(X) or adj(Z) it is the case that I(X, Z | S); list these subsets as $S_1,...,S_r$; list their corresponding p-values using the S function as $p_1,..., p_r$
    b. Find the $p_i$ that maximizes $p_1,...,p_r$.
    c. If $S_i$ does not contain Y, map X—Y—Z to $p_i$ in C.
6. Sort C high to low by p values.
7. For each X—Y—Z in the domain of C
    a. If doing so would not create a bidirected edge,
        i. Orient X->Y<-Z
8. Use the rules in Meek (1995) to orient any further edges that can be oriented in G without create new unshielded colliders.
9. Return G.

Heuristically, we may parallelize steps (3b) and (5). A sensible method of parallelizing (3b) has been proposed; we will just use it; this is the method proposed for the PC-Stable algorithm (Colombo and Maathuis, 2014). The idea is as follows. The adjacency search for PC is divided into "depths." At depth 0, all unconditional independence tests are done. At depth 1, all independence tests are done conditional on sets of size 1. At depth 2, we condition on sets of size 2, and so on. The PC-Stable algorithm changes this a little. Depth 0 is the same. But for depth 1, we simply calculate all of the edges that *would be* removed, but without removing them. So all adjacents of every node are fixed though the entire depth 1 search. Then, at the end, we do all of the removals. This effectively uses the adjacencies of the graph at the end of depth 0 for the depth 1 search. We repeat this for depths 2, 3, etc. For parallelization, this has a remarkable effect, since within each depth all independence judgments are rendered completely independent of each other.

Using the PC-Stable modification has two other advantages. The first is that the adjacency search no longer depends on the order of the variables in the data set. This was the main motivating idea behind the PC-Stable method (Colombo and Maathuis, 2014). Note the collider orientation step of PC-Max is also rendered much more stable in this sense because colliders are added in order of decreasing p-values. Its only risk of instability is that for some X—Y—Z—W, X->Y<-Z and Y->Z<-W might have exactly the same p-value. This may happen, but the event would be very rare. Short of this problem, the orientation phase for PC-Max is stable as well. The second advantage, shown by Colombo and Maathuis and simulation, is that adjacency accuracy for larger models is improved. This is useful for the applications we make of it below.

There is one other step that can readily be parallelized in our proposed algorithm, and that's the collider orientation step itself, step (5). We are calculating a list of colliders that could be added to the graph; this can be done in any order without affecting the outcome. The step of adding the colliders to the graph, step (7), imposes serial constraints, but (5) does not.

## 7. Simulations.

PC-Max scales nicely to large numbers of variables, but to compare it to PC and CPC we need to consider smaller numbers of variables. We compare PC, CPC, PC-Stable, and PC-Max. We include FGS (Ramsey, 2016), a recent implementation of GES (Chickering, 2002; Meek, 1997), a score-based algorithm, using differences of linear, Gaussian BIC scores for independence checks, as explained in section 1, since it has shown superior performance on the same problem. For our comparison, we create random graphs with 1000 nodes, with average degree 2 or 4, formed by listing the variables and then adding random edges in the forward direction. We parameterize these graphs as linear, Gaussian structural equation models and draw 1000 i.i.d. random samples, using a fixed point technique. We repeat this procedure 10 times for each average degree to produce 20 datasets with known true graphs. We then run each of the algorithms on each data set and report average statistics across runs. We use the following formulas for precision and recall:

Precision = TP / (TP + FP)
Recall = TP / (TP + FN)

Here, TP is the count of true positives, FP is the count of false positives (occurring in the estimated graph but not in the true graph) and FN is the count of false negatives (occurring in the true graph but not in the estimated graph). We report "adjacency precision" (AP) which is a precision calculation where a true positive adjacency is one that is in both the pattern of the true graph and the graph estimated by the respective procedure. Similarly, we report "adjacency recall" (AR) We also report "arrowhead precision" (AHP), where a true positive arrowhead is one that is in both the true and estimated graph. Thus, if X->Y in the true pattern and X->Y or X<->Y in the estimated graph, this counts as a true arrowhead; all other arrowheads are false. We also calculate "arrowhead recall" (AHR) using the same information. Finally, we record the percentage of bidirected edges in the graph (BID) and the elapsed time of the algorithm in seconds (E). For algorithms that require an independence test, the Fisher Z test was used, with alpha = 0.001. For FGS, the SEM BIC score was used, with a multiplier on the BIC penalty of 4. Results are reported in Table 1.

Looking at this table, the adjacency numbers are all excellent. Adjacency precisions (AP) are basically perfect for all algorithms; adjacency recalls are just as good for AP = 2 and for FGS at average degree 4; for the rest of the algorithms for average degree 4, they fall to 0.85, still quite good. For orientation there's quite a range of arrow precisions and recalls. For arrowhead precision (AHP), CPC, PC-Max and FGS are in the top tier, with PC and PC-Stable falling behind. For arrowhead recall (AHR) FGS is in a class of its own, followed by PC and PC-stable, though only because one arrow for each bidirected edge is counted correct. Slightly behind that are PC-Max and, further behind, CPC. Bidirected edges are quite telling. With no correction, PC and PC-Stable have large numbers of bidirected edges or models of this size; the rest of the algorithms have essentially none. Timing results are also quite telling. PC and CPC take much longer than any of the other three algorithms at average degree 2, and this is because they're not parallelized. These experiments were carried out on a MacBook Pro with 4 processors, on a 2-core 3.1 GHz Intel Core I7 processor, using 12 G of RAM, so it's not surprising that the non-parallelized processes take about 4 times as long to run as the parallelized processes. At average degree 4, more differences become apparent. In particular, there are two algorithms that are storing sepsets for use in the collider orientation step. For our implementation of PC-Stable, in particular, this ends up costing time. Otherwise, PC and CPC are using the same adjacency search, and PC-Stable and PC-Max are using the same adjacency search. PC and PC-Stable are using the same collider orientation search. All are using the same implementation of the Meek final orientation rules.

| Algorithm | Avg Deg | AP | AR | AHP | AHR | BID | E |
|---|---|---|---|---|---|---|---|
| PC | 2 | 0.98 | 0.96 | 0.77 | 0.93 | 0.08 | 8.74 |
| CPC | 2 | 0.98 | 0.96 | 1.00 | 0.87 | 0.00 | 9.04 |
| PC-Stable | 2 | 0.98 | 0.96 | 0.64 | 0.92 | 0.18 | 1.08 |
| PC-Max | 2 | 0.98 | 0.96 | 0.98 | 0.91 | 0.00 | 1.27 |
| FGS | 2 | 1.00 | 0.98 | 0.99 | 0.97 | 0.00 | 1.29 |
| PC | 4 | 0.99 | 0.85 | 0.66 | 0.86 | 0.28 | 10.55 |
| CPC | 4 | 0.99 | 0.85 | 0.99 | 0.75 | 0.00 | 12.43 |
| PC-Stable | 4 | 1.00 | 0.85 | 0.57 | 0.86 | 0.46 | 39.42 |
| PC-Max | 4 | 1.00 | 0.85 | 0.97 | 0.81 | 0.00 | 4.80 |
| FGS | 4 | 0.97 | 0.94 | 0.96 | 0.93 | 0.00 | 6.22 |

**Table 1. A comparison of PC, CPC, PC-Stable, PC-Max, and FGS for models with graphs with 1000 nodes. "Avg Deg" = average degree of the graph; AP = Adjacency Precision; AR = Adjacency Recall; AHP = Arrowhead Precision; AHR = Arrowhead Recall, BID = Percent Bidirected edges; E is elapsed time in seconds. For independence tests, an alpha of 0.001 was used; for scores, a penalty multiplier of 4 was used.**

What is not shown in the table, for lack of space, is that CPC renders a fair number of unshielded triples in the graph ambiguous. For degree 2, it renders 4% of all unshielded triples ambiguous; for degree 4, it renders 7% of all unshielded triples ambiguous. None of the other algorithms mark ambiguities. In this sense, algorithms such as PC-Max or FGS that do not output ambiguities but have otherwise similar profiles are to be preferred.

The two best algorithms overall are PC-Max and FGS, the first a constraint-based search, the second a scoring search. It is remarkable that they are tied for precision, with PC-Max being slightly behind on recall. On timing, PC-Max is faster, but this may simply be because it doesn't find as many orientations.

To increase to larger numbers of variables, we restrict ourselves to algorithms that run quickly in Table 1, PC-Max and FGS. Since it has been so far been considered a de facto upper limit for PC-style algorithms, we choose to use 20,000 variables with graphs of average degree 2 and 4. Because of the larger numbers of tests done, we lower our alpha level for the Fisher Z test to 0.00001. The multiplier on the BIC penalty is left at 4 for the SEM BIC score. Since neither of these algorithms is capable of orienting a bidirected edge, we remove the BID column from the table.

| Alg | Ave Degree | AP | AR | AHP | AHR | E |
|---|---|---|---|---|---|---|
| PC-Max | 2 | 1.00 | 0.94 | 0.99 | 0.87 | 225.91 |
| FGS | 2 | 1.00 | 0.98 | 1.00 | 0.98 | 175.24 |
| PC-Max | 4 | 1.00 | 0.79 | 0.99 | 0.74 | 335.25 |
| FGS | 4 | 1.00 | 0.93 | 1.00 | 0.93 | 264.50 |

**Table 2. A comparison of PC-Max, and FGS for models with graphs with 20,000 variables. "Avg Deg" = average degree of the graph; AP = Adjacency Precision; AR = Adjacency Recall; AHP = Arrowhead Precision; AHR = Arrowhead Recall, E is elapsed time in seconds. For independence tests, an alpha of 0.00001 was used; for scores, a penalty multiplier of 4 was used.**

In Table 2, precisions are excellent for both algorithms. Recalls are quite good across the board, with recalls for FGS being higher than those for PC-Max. PC-Max is noticeably slower than FGS on the same data.

FGS has been scaled for average degree 2 graphs to 1,000,000 nodes in previous work (Ramsey, 2016); Tables 1 and 2 suggest that PC-Max could be scaled possibly that far as well, with good accuracy, though it would take considerable patience to do so and perhaps a machine with more processors than were used for the FGS example. We may yet try, in the spirit of Mount Everest climbing.

The implementations used for the above procedures were those in the Tetrad Freeware, https://github.com/cmu-phil/tetrad, branch 'development', commit 655fddc.

## 8. Conclusion.

We've suggested a very simple tweak to the PC algorithm, which happily has pleasant side effects. We've merely pointed out that when faced with a choice of conditioning sets, one of which must be for independence, it is sensible to pick the conditioning set that yields the highest p-value under independence. We've applied this reasoning to the collider orientation step in two ways, first, in the identification of colliders, and second, in the choice of which colliders to orient in the graph to avoid bidirected edges. The result is a *constraint-based* algorithm that performs almost as well as the *score-based* algorithm FGS. Previous limitations on the numbers of variables that accurately can be handled by a PC-style algorithm for small samples are easily surpassed in simulation.

It is not lost on us that the max p-value reasoning can be applied to other algorithms. Two other algorithms that we have applied it to, with good success, are FCI (Spirtes et al., 2000) and CCD (Richardson and Spirtes, 1999). FCI relaxes the constraint of causal sufficiency for PC and allows for latent common causes, drawing out the consequences of this for adjacency and orientation. CCD keeps the assumption of causal sufficiency but relaxes the assumption of acyclicity, allowing cyclic models to be discovered. In both cases, we have rendered these algorithms more accurate and much more scalable than otherwise. Hopefully, we will be able to describe the results of those renderings soon.